# DS-SLAM: A Semantic Visual SLAM towards Dynamic Environments

Chao Yu[1], Zuxin Liu[2], Xin-Jun Liu*[1], Fugui Xie[1], Yi Yang[3], Qi Wei[3], Qiao Fei*[3]

*Abstract*— Simultaneous Localization and Mapping (SLAM) is considered to be a fundamental capability for intelligent mobile robots. Over the past decades, many impressed SLAM systems have been developed and achieved good performance under certain circumstances. However, some problems are still not well solved, for example, how to tackle the moving objects in the dynamic environments, how to make the robots truly understand the surroundings and accomplish advanced tasks. In this paper, a robust semantic visual SLAM towards dynamic environments named DS-SLAM is proposed. Five threads run in parallel in DS-SLAM: tracking, semantic segmentation, local mapping, loop closing and dense semantic map creation. DS-SLAM combines semantic segmentation network with moving consistency check method to reduce the impact of dynamic objects, and thus the localization accuracy is highly improved in dynamic environments. Meanwhile, a dense semantic octo-tree map is produced, which could be employed for high-level tasks. We conduct experiments both on TUM RGB-D dataset and in real-world environment. The results demonstrate the absolute trajectory accuracy in DS-SLAM can be improved one order of magnitude compared with ORB-SLAM2. It is one of the state-of-the-art SLAM systems in high-dynamic environments.

## I. INTRODUCTION

In the past years, visual SLAM has been extensively investigated, because images store a wealth of information and can be employed for other vision-based applications, like semantic segmentation and object detection. The framework of modern visual SLAM system is quite mature, which often consists of several essential parts, such as feature extraction front-end, state estimation back-end, loop closure detection and so forth. Besides, some advanced SLAM algorithms have already attained satisfactory performance, such as ORB-SLAM2[1], LSD-SLAM[2].

However, some issues remain unsolved, for example, most of the existing algorithms are fragile. On the one hand, they have difficulty in handling all kinds of environments, such as extraordinarily dynamic or rough environments. On the other hand, their map models are often based on geometric information, like the landmark-based map and point cloud map, so they do not provide any high-level understanding of the surroundings. According to [3], SLAM enters the robust-perception age and more research to achieve genuinely robust perception and navigation for autonomous robots is needed.

The typical SLAM methods only provide a map with geometric entities (points, planes, etc.), which do not have semantic attributes distinction between them. However, the semantic information is needed for the robot to understand the scene surrounding them. With the development of deep learning, some networks could achieve good performance in semantic segmentation[4]. Therefore, combining these networks with SLAM could produce the semantic map, and thus improve the perception level of robots [5, 6].

The robustness of the SLAM system in dynamic environments is also a challenge. While modern SLAM system has been successfully demonstrated mostly in specific circumstances, unexpected changes of surroundings would probably corrupt the quality of the state estimation and even lead to system failure. For example, the presence of dynamics in the environment, like walking people, might deceive feature association in vision-based SLAM system. Although some progress has been made to reduce the impact of the changes of the environment in laser-based SLAM, for instance, some approaches combine inertial measurement unit with visual SLAM to improve robustness, like [7], but the problem is still not well solved in pure vision-based SLAM.

In this paper, we focus on reducing the impact of dynamic objects in vision-based SLAM by combining semantic segmentation network with optical flow method and meanwhile providing a semantic presentation of the octo-tree map [8], which could be employed for high-level tasks for robots. The overview of DS-SLAM is shown in Fig. 1.

The main contributions of this paper include:

1. A complete semantic SLAM system in dynamic environments (DS-SLAM) is proposed based on ORB-SLAM2 [2], which could reduce the influence of dynamic objects on pose estimation. The effectiveness of the system is evaluated on TUM RGB-D dataset [9]. The results indicate that DS-SLAM outperforms ORB-SLAM2 significantly regarding accuracy and robustness in dynamic environments. The system is also integrated with Robot Operating System (ROS) [10], and its performance is verified by testing DS-SLAM on a robot in a real environment.

2. We put a real-time semantic segmentation network in an independent thread, which combining semantic segmentation with moving consistency check method to filter out dynamic portions of the scene, like walking people. Thus the performance of localization module and mapping module is

Resrach supported by the National Natural Science Foundation of China under Grant 91648116 and 51425501.

[1] Department of Mechanical Engineering, Tsinghua University, Beijing, China
[2] School of Instrumentation Science and Opto-electronics Engineering, Beihang University, Beijing, China
[3] Department of Electronic Engineering, Tsinghua University, Beijing, China
*corresponding author: xinjunliu@mail.tsinghua.edu.cn and qiaofei@mail.tsinghua.edu.cn

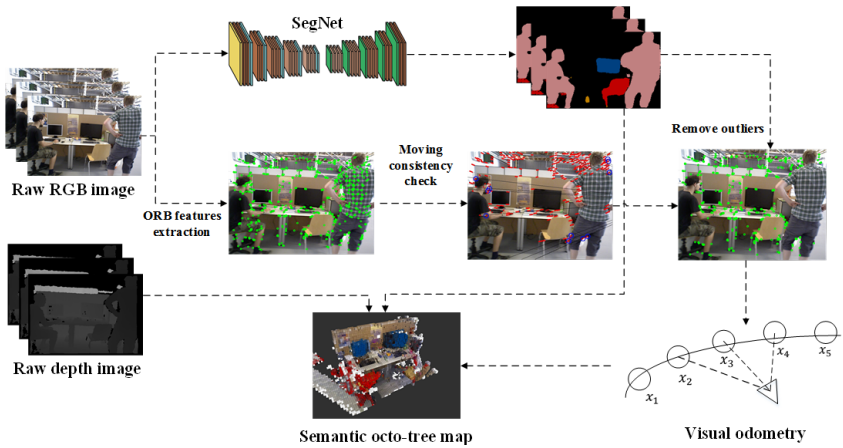

Figure 1. The overview of DS-SLAM. The raw RGB image is utilized to semantic segmentation and moving consistency check simultaneously. Then remove outliers and estimate pose. Semantic octo-tree map is built in an independent thread based on the pose, depth image, and semantic segmentation results.

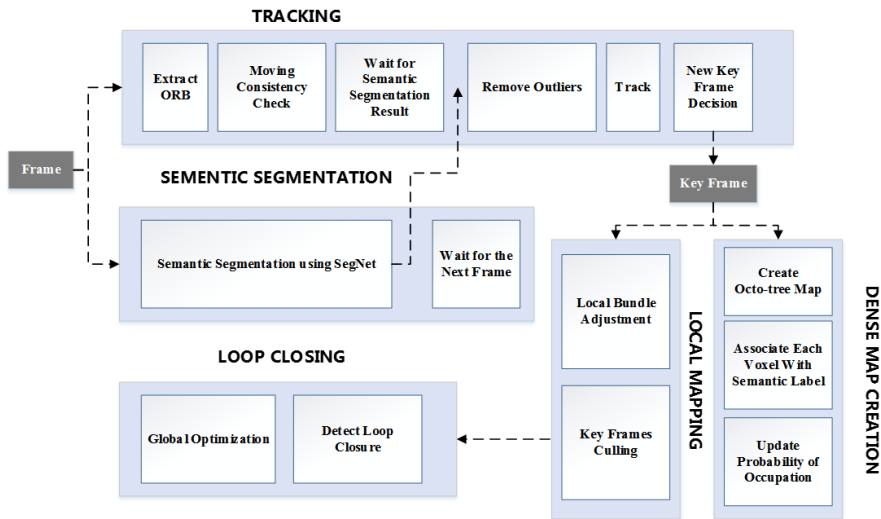

Figure 2. The framework of DS-SLAM. Local mapping thread and loop closing thread are the same as ORB-SLAM2. The former one processes new keyframes and performs local bundle adjustment to achieve an optimal reconstruction in the surroundings of the camera pose, while the later one searches for loops and perform a graph optimization if a loop is detected.

improved in respects of robustness and accuracy in dynamic scenarios.

3. DS-SLAM creates a separate thread to build a dense semantic 3D octo-tree map [8]. The dense semantic 3D octo-tree map adopts log-odds score method to filter out unstable voxels and update these voxels' semantic meaning.

In the rest of this paper, the structure is as follows. Section II provides an overview of various current accomplishments in respects of semantic SLAM and SLAM in dynamic environments. Then section III presents the framework of this whole SLAM system in detail, that how to detect dynamic objects and produce semantic maps. Subsequently, Section IV provides qualitative and quantitative results of performance of DS-SLAM both on TUM RGB-D dataset [9] and in a real-world environment to demonstrate the effectiveness and accuracy of the system. Finally, a brief conclusion and discussion are given in section V.

## II. RELATED WORKS

### A. Semantic SLAM

Since purely geometry-based maps could not provide conceptual knowledge of the surroundings to facilitate complex tasks, associating semantic concepts with geometry entities in the environment has become a popular research domain recently. In many previous works, the semantic maps often consisted of two parts: a geometric part and a semantic part. Afterwards, a group of approaches trained the object recognition subsystem in advance and attached semantic information to the object models recognized. Vasudevan et al. [11] apply a simple naïve Bayesian classifier to identify the scene categories and give a hierarchical probabilistic representation of space based on pre-trained objects. More recently, an efficient description of 3D models was proposed by Sengupta et al. [12], which embedded an octo-tree into a hierarchical robust Markov Random Filed to provide each voxel in the map with the semantic label. With the advent of deep learning, the results of semantic segmentation have been vastly improved. Sünderhauf et al. [13] apply convolutional neural network to 3D points segmentation and correlate semantic label with object based on the nearest neighbor method, and then add or update the map of target object point cloud information and subordinate type confidence value.

However, their work only focuses on semantic mapping and object recognition, while the semantic information is not well used in other parts. Bao et al. [14] first try to estimate camera pose, points, and objects jointly utilizing both geometry and semantic attributes in the scene, which significantly improve the object recognition accuracy. In this paper, semantic information is not only used to generate an octo-tree map based representation of the environment but also utilized to filter outliers in the process of tracking in a dynamic environment. To perform semantic segmentation accurately, we use semantic segmentation network to follow the state-of-the-art line of work.

### B. SLAM in dynamic environments

A basic assumption in most current SLAM approaches is that the environment is static. However, active objects like humans, exist in many real-world scenes. Therefore, most state-of-the-art approaches that initially designed for doing SLAM in static environments are not capable of handling severe dynamic scenarios.

To address this issue, identifying the dynamic objects from the static parts, and then discarding them before pose estimation is necessary. In dense SLAM, many moving target detection methods often based on optical flow technique. Optical flow is generated if movements exist in the image, so static background and moving target could be distinguished by computing the inconsistency of optical flow. For example, Fang et al. [15] use optimum-estimation and uniform sampling method to detect dynamic objects. Although it is efficient in time comparing with other kinds of derivative optical flow methods, its accuracy is lower, and the computation load is still large, especially when sampling densely in a big image. Wang et al. [16] detect dynamic objects by clustering the image based on point trajectories and exclude them from energy function minimization. Their method is robust, but cannot be performed in real-time.

## III. SYSTEM INTRODUCTION

In this section, DS-SLAM will be introduced in details. This section contains five aspects. First, architecture chart of the framework of DS-SLAM is presented. Second, we give a brief introduction to the real-time semantic segmentation method that adopted in DS-SLAM. Then the algorithm that is used for checking the moving consistency of the feature points is introduced. Subsequently, the outliers rejection method is demonstrated, which combine the semantic segmentation and the moving consistency check to filter out dynamic objects. Finally, we present the method to build semantic octo-tree map.

### A. Framework of DS-SLAM

In real-world applications, accurate pose estimation and reliability in harsh environments are critical factors to evaluate autonomous robots. ORB-SLAM2 has an excellent performance in most practical situations. Therefore, ORB-SLAM2 is adopted in DS-SLAM to provide a global feature-based SLAM solution that enables us to detect dynamic objects and produce semantic octo-tree map. The overview of DS-SLAM is shown in Fig. 1.

Five threads run in parallel in DS-SLAM: tracking, semantic segmentation, local mapping, loop closing and dense map creation. The framework of the system is shown in Fig. 2. The raw RGB images captured by Kinect2 are processed in tracking thread and semantic segmentation thread simultaneously. The tracking thread first extracts ORB feature points, then check moving consistency of the feature points roughly and save the potential outliers. Then the tracking thread waits for the image that has pixel-wise semantic label predicted by semantic segmentation thread. After the segmentation result arrives, the ORB feature points outliers located in moving objects will be discarded based on the result and the potential outliers detected before. Then, the transformation matrix is calculated by matching the rest of stable feature points.

### B. Semantic Segmentation

This system is designed for real-world applications, so there should be a balance between real-time and accuracy. DS-SLAM adopts SegNet [4] to provide pixel-wise semantic segmentation based on caffe [17] in real-time. The SegNet trained on PASCAL VOC dataset [18] could segment 20 classes in total. In real applications, people are most likely to be dynamic objects, so we assume that feature points located in people are most likely to be outliers.

### C. Moving Consistency Check

Since motion segmentation is time-consuming and semantic segmentation results can be obtained from another thread, we only need to determine whether the key points in segmentation results are moving. If some points are determined to be dynamic within a segmented object, then the object could be regarded as a dynamic object. The idea of moving points detection in this paper is straightforward. The first step is to calculate optical flow pyramid to get the matched feature points in the current frame. Then if the matched pair is too close to the edge of the image or the pixel difference of the $3 \times 3$ image block at the center of the matched pair is too large, the matched pair will be discarded. The third step is to find fundamental matrix by using RANSAC with the most inliers. Then calculate the epipolar line in the current frame by using the fundamental matrix. Finally, determine whether the distance from a matched point to its corresponding epipolar line is less than a certain threshold. If the distance is greater than the threshold, then the matched point will be determined to be moving.

The fundamental matrix maps the feature points in the last frame to the corresponding search domain in the current frame, that is, the epipolar line. Let $p_1, p_2$ denote the matched points in the last frame and current frame respectively, and $P_1, P_2$ are their homogeneous coordinate form:

$$P_1 = [u_1, v_1, 1], P_2 = [u_2, v_2, 1], \\ p_1 = [u_1, v_1], p_2 = [u_2, v_2] \quad (1)$$

where $u, v$ are the values in the image frame. Then the epipolar line that is denoted as $I_1$ can be computed by the following equation:

$$I_1 = \begin{bmatrix} X \\ Y \\ Z \end{bmatrix} = FP_1 = F \begin{bmatrix} u_1 \\ v_1 \\ 1 \end{bmatrix} \quad (2)$$

where $X, Y, Z$ represent line vector and $F$ represents fundamental matrix. Then the distance from the matched point to its corresponding epipolar line is determined as follows:

$$D = \frac{|P_2^T F P_1|}{\sqrt{\|X\|^2 + \|Y\|^2}} \quad (3)$$

where $D$ represents the distance. The algorithm to check moving consistency and determine the dynamic points is shown in Algorithm 1, where $\varepsilon$ is a preset threshold value.

**Algorithm 1** Dynamic Points Detection Algorithm
**Input:** Previous frame, $F_1$; Previous frame's feature points, $P_1$; Current frame, $F_2$;
**Output:** The set of dynamic points, $S$;
1: Current frame's feature points $P_2 = CalcOpticalFlowPyrLK(F_1, F_2, P_1)$
2: Remove outliers in $P_2$
3: $FM = FindFundamentalMatrix(P_1, P_2)$
4: **for** each matched pairs $p_1, p_2$ in $P_1, P_2$ **do**
5:    $I_1 = FindEpipolarLine(p_1, F_M)$
6:    $D = CalcDistanceFromEpipolarLine(p_2, I_1)$
7:    **if** $D > \epsilon$ **then**
8:       Append $p_2$ to $S$
9:    **end if**
10: **end for**

### D. Outliers Rejection

Because of the flexible deformation and complex motion of moving objects, such as human, it is difficult for moving consistency check method to extract the contours of the complete dynamic regions, and the time cost of extracting the whole contours is very expensive. In DS-SLAM, since the semantic segmentation network have been employed, the full outlines of objects could be easily obtained. Our idea is to combine semantic information and moving consistency check results to complete the establishment of the two-level semantic knowledge base: the object is moving or not moving. If there are certain number of dynamic points produced by moving consistency check fall in the contours of a segmented object, then this object is determined to be moving. If the segmented object is determined to be moving, then remove all the feature points located in the object's contour. In this way, outliers can be eliminated precisely. Besides, the impact of wrong segmentation can also be reduced to a certain extent.

Furthermore, the time that the tracking thread waits for the semantic segmentation result from another thread could be taken full use of. During the waiting period, moving consistency check could be performed. In the next experimental results section, Table IV gives the consumption time in DS-SLAM. It can be seen that the time used in the tracking thread is approximately equal to the semantic segmentation thread.

Because human activities interfere with robots' localization seriously in most real application scenarios, and the categories that semantic segmentation networks can identify are limited, we would take the human as a typical representative of dynamic objects in the rest of this paper. In theory DS-SLAM applies to any multiple recognized and segmented dynamic objects.

After the semantic segmentation result come out, if no people are detected, then all the ORB features would be matched with the last frame directly to predict the pose. Otherwise, determine whether the people are moving by using the moving consistency check results. If the people are determined to be static, then predict the pose directly, otherwise remove all the ORB feature points that fall within the outline of people before matching. In this way, the impact of dynamic objects could be reduced remarkably.

### E. Dense semantic 3D octo-tree map building

The semantic octo-tree mapping thread gets new keyframes from tracking thread and segmentation results from semantic segmentation thread. The keyframes' transform matrix and the depth images are used to generate the local point cloud. Then the local point cloud would be converted and maintained in a global octo-tree map. We adopt the octo-tree representation [8] because it is flexible, compact and updatable. The octo-tree map is stored efficiently, and it is easy to be employed for navigation.

Semantic information is also incorporated into the octo-tree map. Each voxel in the octo-tree map is associated with a specific color, and each color represents a semantic label. For example, the voxel in red color represents it belongs to a sofa, while the voxel in pink represents it belongs to a person. All the processes of modeling and semantic fusion are done in a probabilistic fashion, so it is convenient to update the voxel's attribute. In this way, the dense semantic 3D octo-tree map can provide a basis for mobile robots to accomplish advanced tasks.

DS-SLAM is oriented to deal with dynamic environments, so the dynamic objects should not exist in the map. Semantic segmentation results can help us filter out dynamic objects effectively. However, the accuracy of semantic segmentation is limited. In complex situations, for example, objects overlap each other, the semantic segmentation results may be incomplete or even wrong. To address this problem, log-odds score method [8] is used in DS-SLAM to minimize the impact of dynamic objects. The log odds score is used to represent the possibility of an individual voxel being occupied quantitatively. Let $p \in [0,1]$ denote the probability of a voxel being occupied, and $l \in R$ is denoted as the log odds score of the probability. $l$ can be calculated by logit transform:

$$l = \text{logit}(p) = \log(\frac{p}{1-p}) \quad (4)$$

The inverse transform:

$$p = \text{logit}^{-1}(l) = \frac{\exp(l)}{\exp(l)+1} \quad (5)$$

Let $Z_t$ denote the observation result of a voxel $n$ at time $t$, and its log odds score from the beginning to time $t$ is $L(n|Z_{1:t})$, Then at time $t+1$, the log odds score of voxel $n$ can be calculated as follows:

$$L(n|z_{1:t+1}) = L(n|z_{1:t-1}) + L(n|z_t) \quad (6)$$

where $L(n|Z_t)$ is equal to $\tau$ if the voxel $n$ is observed to be occupied at time $t$, otherwise 0. The increment $\tau$ is a pre-defined value. This formula represents that the log odds score of a voxel will increase when the voxel is repeatedly observed to be occupied, otherwise decrease. The occupied probability $p$ of a voxel can be calculated by inverse logit transform. Only when the occupying probability $p$ is greater than a pre-defined threshold, the voxel is regarded to be occupied and will be visualized in the octo-tree map. In other words, the voxels that have been observed to be occupied many times are considered to be stable occupied voxels. With this method, we can handle the map building problem in dynamic environments well.

## IV. Experimental results

In this section, experimental results would be presented to demonstrate the effectiveness of DS-SLAM. We tested the time required for each module, and evaluate the performance of DS-SLAM in dynamic environments with public TUM RGB-D dataset. The sequences in TUM RGB-D dataset containing dynamic objects are challenging because the moving people in the observation could corrupt the robustness and accuracy of SLAM algorithm. In extreme case, the moving people occupy more than half of the image. We also integrate DS-SLAM with ROS and qualitatively demonstrate it on a physical robot in typical dynamic environments to evaluate its real-time performance and dense semantic 3D octo-tree map creation module. All the experiments are performed on a computer with Intel i7 CPU, P4000 GPU, and 32GB memory. The physical robot is TurtleBot2, and image sequences are captured by Kinect V2 camera.

### A. Evaluation using TUM RGB-D dataset

The TUM RGB-D dataset [9] provides several sequences in dynamic environments with accurate ground truth obtained with an external motion capture system, such as walking, sitting and desk. The walking sequences are mainly used for our experiments because some people walk back and forth in the walking sequences. The people in these sequences could be regarded as high-dynamic objects, and they are most difficult to handle with. The sitting sequences are also used, but they are considered as low-dynamic sequences as the persons in them just move a little bit.

DS-SLAM adopt ORB-SLAM2 [1] as global SLAM solution, which it is recognized as one of the most outstanding and stable SLAM algorithms at present, so we make a comparison between them. Metrics Absolute Trajectory Error (ATE) and Relative Pose Error (RPE) are used for quantitative evaluation. Metric ATE stands for global consistency of trajectory, while metric RPE measures the translational and rotational drift [9].

The quantitative comparison results are shown in Table I - Table III, where xyz, static, rpy and half in the first column stand for four types of camera ego-motions [9], for example, xyz represents the camera moves along the x-y-z axes. We present the values of RMSE, Mean Error, Median Error and Standard Deviation (S.D.) in this paper, while RMSE and S.D. are more concerned because they can better indicate the robustness and stability of the system. We also show the values of improvement of DS-SLAM compared to the original ORB-SLAM2. The improvement values in the tables are calculated as follows:

$$\eta = \frac{o - r}{o} \times 100\% \qquad (7)$$

where $\eta$ represents the value of improvement, $o$ represents the value of ORB-SLAM2 and $r$ represents the value of DS-SLAM.

As we can see from Table I - Table III, DS-SLAM can make the performance in most high-dynamic sequences get an order of magnitude improvement. In terms of ATE, the RMSE and S.D. improvement values can reach up to 97.91% and 97.94% respectively. The results indicate that DS-SLAM can improve the robustness and stability of SLAM system in high-dynamic environments significantly. However, in low-dynamic sequences, for example, the fr3_sitting_static sequence, the improvements of performance are not obvious. We think the reason is that original ORB-SLAM2 can handle the low-dynamic scenarios easily and achieve good performance, so the space that can be improved is limited.

Fig. 3. and Fig. 4. show selected ATE and RPE plots from ORB-SLAM2 and DS-SLAM in high dynamic fr3_walking_xyz sequence respectively. As we can see, the errors are significantly reduced in DS-SLAM.

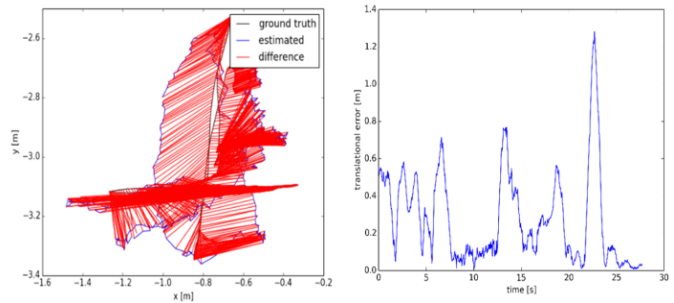

Figure 3.    ATE and RPE from ORB-SLAM2

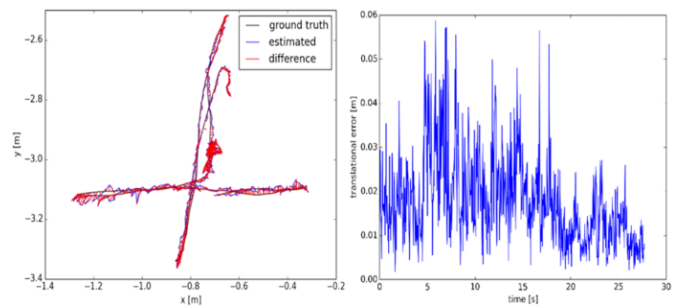

Figure 4.    ATE and RPE from DS-SLAM

For practical applications, real-time performance is a crucial indicator to evaluate SLAM system. We test the time required for some major modules to process. The results are shown in Table IV. The average time in the main thread to process each frame is 59.4ms, including semantic segmentation, visual odometry estimation, pose graph optimization and dense 3D semantic octo-tree map creation. Compared with previous non real-time methods to filter out dynamic objects, such as [19], DS-SLAM is more satisfied with the needs of the real time.

TABLE I. RESULTS OF METRIC ROTATIONAL DRIFT (RPE)

| Sequences | ORB-SLAM2 | | | | DS-SLAM | | | | Improvements | | | |
|---|---|---|---|---|---|---|---|---|---|---|---|---|
| | RMSE | Mean | Median | S.D. | RMSE | Mean | Median | S.D. | RMSE | Mean | Median | S.D. |
| fr3_walking_xyz | 7.7432 | 5.8765 | 4.5340 | 4.9895 | 0.8266 | 0.5836 | 0.4192 | 0.5826 | 89.32% | 90.07% | 90.75% | 88.32% |
| fr3_walking_static | 3.8958 | 1.6845 | 0.3571 | 3.5095 | 0.2690 | 0.2416 | 0.2259 | 0.1182 | 93.09% | 85.66% | 36.75% | 96.63% |
| fr3_walking_rpy | 8.0802 | 5.4558 | 2.7828 | 5.9499 | 3.0042 | 1.9187 | 0.9902 | 2.3065 | 62.82% | 64.83% | 64.42% | 61.23% |
| fr3_walking_half | 7.3744 | 4.5917 | 1.8143 | 5.7558 | 0.8142 | 0.7033 | 0.6217 | 0.4101 | 88.96% | 84.68% | 65.73% | 92.87% |
| fr3_sitting_static | 0.2881 | 0.2598 | 0.2493 | 0.1244 | 0.2735 | 0.2450 | 0.2351 | 0.1215 | 5.07% | 5.70% | 5.68% | 2.36% |

TABLE II. RESULTS OF METRIC TRANSLATIONAL DRIFT (RPE)

| Sequences | ORB-SLAM2 | | | | DS-SLAM | | | | Improvements | | | |
|---|---|---|---|---|---|---|---|---|---|---|---|---|
| | RMSE | Mean | Median | S.D. | RMSE | Mean | Median | S.D. | RMSE | Mean | Median | S.D. |
| fr3_walking_xyz | 0.4124 | 0.3110 | 0.2465 | 0.2684 | 0.0333 | 0.0238 | 0.0181 | 0.0229 | 91.93% | 92.34% | 92.66% | 91.48% |
| fr3_walking_static | 0.2162 | 0.0905 | 0.0155 | 0.1962 | 0.0102 | 0.0091 | 0.0082 | 0.0048 | 95.27% | 90.00% | 47.07% | 97.58% |
| fr3_walking_rpy | 0.4249 | 0.2825 | 0.1487 | 0.3166 | 0.1503 | 0.0942 | 0.0457 | 0.1168 | 64.64% | 66.66% | 69.24% | 63.10% |
| fr3_walking_half | 0.3550 | 0.2161 | 0.0774 | 0.2810 | 0.0297 | 0.0256 | 0.0226 | 0.0152 | 91.62% | 88.16% | 70.74% | 94.60% |
| fr3_sitting_static | 0.0095 | 0.0083 | 0.0073 | 0.0046 | 0.0078 | 0.0068 | 0.0061 | 0.0038 | 17.61% | 17.81% | 17.01% | 16.96% |

TABLE III. RESULTS OF METRICS ABSOLUTE TRAJECTORY ERROR (ATE)

| Sequences | ORB-SLAM2 | | | | DS-SLAM | | | | Improvements | | | |
|---|---|---|---|---|---|---|---|---|---|---|---|---|
| | RMSE | Mean | Median | S.D. | RMSE | Mean | Median | S.D. | RMSE | Mean | Median | S.D. |
| fr3_walking_xyz | 0.7521 | 0.6492 | 0.5857 | 0.3759 | 0.0247 | 0.0186 | 0.0151 | 0.0161 | 96.71% | 97.13% | 97.42% | 95.71% |
| fr3_walking_static | 0.3900 | 0.3554 | 0.3087 | 0.1602 | 0.0081 | 0.0073 | 0.0067 | 0.0036 | 97.91% | 97.95% | 97.82% | 97.74% |
| fr3_walking_rpy | 0.8705 | 0.7425 | 0.7059 | 0.4520 | 0.4442 | 0.3768 | 0.2835 | 0.2350 | 48.97% | 49.26% | 59.84% | 48.02% |
| fr3_walking_half | 0.4863 | 0.4272 | 0.3964 | 0.2290 | 0.0303 | 0.0258 | 0.0222 | 0.0159 | 93.76% | 93.95% | 94.40% | 93.05% |
| fr3_sitting_static | 0.0087 | 0.0076 | 0.0066 | 0.0043 | 0.0065 | 0.0055 | 0.0049 | 0.0033 | 25.94% | 26.87% | 26.29% | 23.15% |

TABLE IV. TIME EVALUATION

| Module | ORB feature extraction | Moving consistency check | Semantic segmentation |
|---|---|---|---|
| Thread | Tracking | Tracking | Semantic segmentation |
| Time(ms) | 9.375046 | 29.50869 | 37.57330 |

### B. Evaluation in Real Environment

To demonstrate the robustness and real-time performance of DS-SLAM, we integrate it with ROS and conduct extensive experiments in a laboratory environment. Images are captured by Kinect V2 camera with $960 \times 540$ resolution.

Fig. 5. qualitatively demonstrate the results of outliers rejection. The sub-figures from top row to bottom row are ORB feature extraction results, optical flow based moving consistency check results, semantic segmentation results, and images after outliers removal respectively. The green dots represent the location of ORB feature points, and the red short lines represent the direction of optical flow. As we can see, the person is determined to be moving by moving consistency check, and then feature points that fall in the moving person area are removed effectively. This figure is better viewed in color.

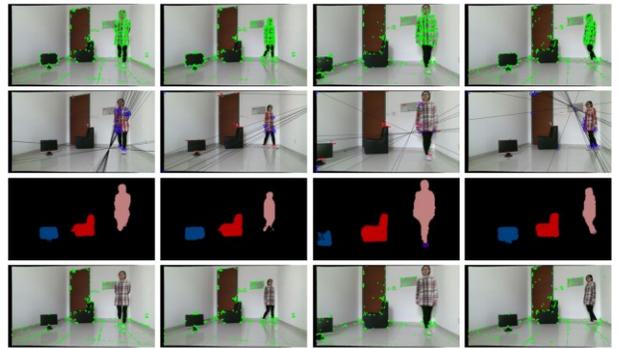

Figure 5. Experimental results in lab environment

Fig. 6. shows the dense 3D semantic octo-tree map reconstruction result. The red voxels represent sofa, and the blue voxels represent monitor. Since log-odds score method is used to filter out unstable voxel in the map, the result of reconstruction is not affected by moving people. Besides, 2D cost map is produced by octo-tree map and can be employed for navigation.

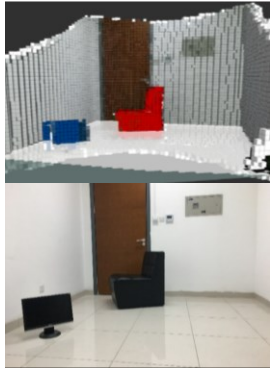

Figure 6.  Dense 3D semantic octo-tree map

## V. Conclusions

In this paper, a complete real-time robust semantic SLAM (DS-SLAM) system is proposed, which could reduce the influence of dynamic objects on pose estimation, and meanwhile provide semantic presentation of the octo-tree map [8]. Five threads run in parallel in DS-SLAM: tracking, semantic segmentation, local mapping, loop closing and dense map creation. A real-time semantic segmentation network SegNet is combined with moving consistency check to filter out dynamic portions of the scene, like walking people. Then the matched feature points would be removed out of those detected dynamic regions, and thus improve the performance of robustness and accuracy in dynamic scenarios. Also, the dense semantic 3D octo-tree map adopts log-odds score method to filter out unstable voxels, and it can be employed for navigation and complex tasks for robots. The effectiveness of the system against dynamic objects is tested on challenging dynamic sequences from TUM RGB-D dataset. The results indicate that DS-SLAM outperforms ORB-SLAM2 significantly regarding accuracy and robustness in the high-dynamic environment. The system is also integrated with Robot Operating System (ROS), and its performance is verified by testing DS-SLAM on a physical robot in the real environment.

However, DS-SLAM still exists some ongoing works. For instance, the types of objects that can be recognized in semantic segmentation network are restricted, which limit its scope of application. The octo-tree map has to be re-built when the loop closure is detected. In the future, we would improve the performance of DS-SLAM in terms of real time by optimizing the moving consistency check method. Moreover, the dense semantic octo-tree map would be adopted for the mobile robots to accomplish high-level tasks.

## Acknowledgment

This work was supported in part by the National Natural Science Foundation of China under Grant 91648116 and 51425501.